\documentclass{article}
\usepackage{spconf,amsmath,graphicx}

\title{FN-Net: Remove the Outliers by Filtering the Noise}
\name{Kai Lv}
\address{Department of Electronic Engineering, Graduate School at Shenzhen, Tsinghua University}
\begin{document}
%
\maketitle
\begin{abstract}
Establishing the correspondence between two images is an important research direction of computer vision. When estimating the relationship between two images, it is often disturbed by outliers. In this paper, we propose a convolutional neural network that can filter the noise of outliers. It can output the probability that the pair of feature points is an inlier and regress the essential matrix representing the relative pose of the camera. The outliers are mainly caused by the noise introduced by the previous processing. The outliers rejection can be treated as a problem of noise elimination, and the soft threshold function has a very good effect on noise reduction. Therefore, we designed an adaptive denoising module based on a soft threshold function to remove noise components in the outliers, to reduce the probability that the outlier is predicted to be an inlier. Experimental results on the YFCC100M dataset show that our method exceeds the state-of-the-art in relative pose estimation.
\end{abstract}
\begin{keywords}
Outlier rejection, noise filtering, soft threshold
\end{keywords}
\section{Introduction}
\label{sec:intro}

Two-view geometry estimation plays an important role in computer vision, which is widely applied to Structure from Motion(SfM)~\cite{16} and visual Simultaneous Localization and Mapping(SLAM)~\cite{10}. Many tasks related to image matching require feature extraction and matching. Due to the limitations of existing feature extraction and match algorithms, the matching results often contain mismatched feature point pairs, which are generally referred to as outliers. Correspondingly, pairs of feature points that match accurately are called inliers. In order to estimate the accurate camera pose, the inlier rate should be as high as possible, so the outliers rejection algorithm is needed to reduce the number of outliers and improve the inlier rate.

In recent years, more and more researchers begin to study the learning-based outlier rejection method~\cite{14}, and many works have achieved good results, especially the PointCN series~\cite{21}. But the PointCN series only considers the contextual information between the inputs and ignores the noise contained in the inputs, which is likely to affect the results. 

The commonly used function for signal denoising is the soft threshold function, which can be formulated as:
$$o=\left\{
\begin{array}{rcl}
i-t,& &{i>t}\\
0, & &{-t\le i\le t}\\ 
i+t,& &{i<t}\\
\end{array} \right.$$
where i represents the input signal, o is the output signal, and t indicates the set threshold value.

The soft threshold function sets the noise which is small relative to the normal signal to zero and preserves the relative amplitude of the normal signal. It can effectively remove the noise component in the signal and can retain the useful signal. However, it is difficult to determine the threshold value, and a wealth of relevant experiences is needed to find an appropriate value. The DRSN~\cite{25} has successfully applied the soft threshold function in the convolutional neural network for fault detection. And it also can be applied in the network to reduce the influence of noise in the outliers and estimate a more accurate camera pose.

When designing an outlier rejection algorithm, we should pay attention to the problem that the input pairs of feature points are generally unordered. When the input pairs of feature points are entered in a different order, the result should be consistent except for the order, that is, the processing is independent of the order of the inputs. For example, use PointNet-like structure~\cite{12} and Context Normalization(CN)~\cite{18} to process the input, and the related network processing results need to be independent of the input order. In addition to the unorder of the inputs, another problem of the sparse matching is that the correspondence cannot find well-defined neighbors~\cite{8} between the inputs like the 3D point cloud, which may require modeling through a bilateral domain model or a graphical model. Because of these two constraints, the operation in the existing network keeps the relative order of the original data and adopts the way of Graph Neural Network(GNN)~\cite{26} to establish the relationship between the spare matchings, which has obtained good performance.

We take the OANet~\cite{22} using CN and GNN as our baseline network and propose a convolutional neural network based on a soft threshold that can filter the noise of the outliers, which we refer to as FN-Net. Its network architecture is shown in Fig. 1.

\begin{figure}[htb]
\begin{minipage}[b]{1.0\linewidth}
 \centerline{\includegraphics[width=9cm]{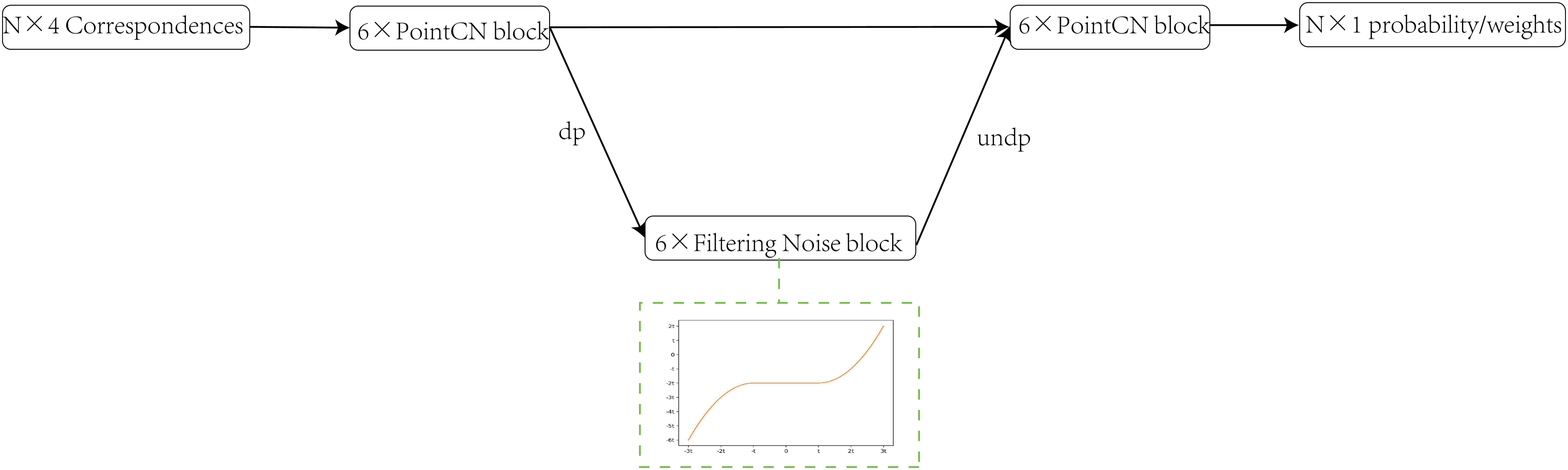}}
\end{minipage}

\caption{The network structure of the FN-Net. dp and undp represent the same differentiable pooling and differentiable unpooling as the OANet. In addition, we propose a noise filtering module to reduce the influence of the noise in the outliers on the results.}
\label{fig:short}
\end{figure}

Our contributions are concentrated on two aspects:

1. By reducing the noise of the outliers, the relative probability of the outlier being predicted as an inlier is lower, the relative probability of the inliers being predicted as an inlier is higher, and the prediction probability distance between inlier and outlier is further.

2. A filter-noise network based on soft threshold function combined with CN structure is designed, which can not only filter the noise in the outliers but also preserve the contextual information between the inputs.

\section{Related Work}
\label{sec:format}

\subsection{Outlier Rejection }
In general, the correspondences after feature extraction and feature matching contain many outliers. The outliers can interfere with the calculation of the camera's pose. RANSAC~\cite{6} is the most commonly used method in various applications, which takes the points that do not conform to the best model as the outliers in an iterative way. USAC~\cite{13}integrates the improvements to RANSAC into a universal framework, making RANSAC more convenient to use. GMS~\cite{bian2017gms} determines whether the match is correct by counting the number of matching correspondence in the neighborhood.

Based on the learning approach, DSAC~\cite{3} is a differentiable network with reference to RANSAC. PointCN~\cite{21} models outlier rejection as a dichotomy problem and an essential matrix regression problem~\cite{21}. It is combined with the weighted eight-point algorithm~\cite{1} to directly regress the essential matrix. It puts forward the CN module has greatly improved the effect. DFE~\cite{14} modified loss function and iterative method. The ${N^3}$ network~\cite{11} enhances PointCN with the soft K nearest neighbor layer. OANet~\cite{22} models the correspondences into graph models and introduces the DiffPool layer and DiffUnpool layer to divide the inputs into different clusters, and after merging the context between the correspondences, the network outputs more accurate weights. Referring to the network structure of the OANet, we replace the level-2 layer of OANet with a module that can filter the noise component of the outliers.

\subsection{Deep CNN Denoisers}
With the rapid development of deep learning, convolutional neural networks are used in the field of image denoising. Burger~\cite{4} was first successful in removing Gaussian noise from images. Later, CSF~\cite{15} and TNRD~\cite{5} integrated the optimization into the denoising network. Zhang~\cite{23} combines residual learning and batch normalization, which achieves a better result than the traditional non-CNN algorithm.  Noise2Noise~\cite{7} obtain better performance without clean data. Red30, MemNet~\cite{17}, BM3DNet~\cite{20}, FFDNet~\cite{24} all provide good ideas for the field of noise removal. The denoising network can also be applied to fault diagnosis. It proposes a deep residual shrinkage network~\cite{25}, which uses a soft threshold function commonly used in signal denoising algorithms to filter the data irrelevant to the current task. The feature of the noise is set to zero to filter out noise. Outliers can be regarded as the result of the superposition of the inliers and the noises, and the problem of outlier rejection can be treated as a kind of denoising problem.

\section{Method}
\label{sec:pagestyle}

\subsection{Problem Formulation}
In order to get the relative pose between the cameras that take the paired images, we need to remove the outliers. The general process is to first extract features from two images, then they are matched by a matching algorithm, such as k nearest neighbor. Since these two steps only rely on local features, many mismatched correspondences will be produced. After removing the outliers, the essential matrix can be calculated, which represents the relative pose of the cameras.

The input of the outlier rejection algorithm is generally a set of feature point pairs(fpp)$\in R^{N\times 4}$:
\begin{equation}
fpp=[p_{11},p_{21},p_{12},p_{22},...,p_{1N},p_{2N}],p_{ij}=[p_x,p_y],
\end{equation}
where N is the number of the feature point pair, $p_{ij}$ denotes the coordinates of the jth feature points in the ith image, $p_x$, $p_y$ represents the coordinate value of the pixel plane.

According to the different methods of calculating the essential matrix, the output of the outlier rejection algorithm is the label or the probability that the input point pair is an inlier.
\begin{equation}
lable/pro=pm(fpp),
\end{equation}
where pm($\cdot$) refers to the process method of obtaining the predicted label or weight values of the input correspondences, lable is the inlier or outlier label correspond to each correspondence, 0 represents the outlier, and 1 denotes the inlier, pro is the probability of each correspondence to become an inlier with the scope in [0,1), which is the weight of calculating the essential matrix.

In general, the output probability is easier to estimate the exact essential matrix than the output label. And the weighted eight-point method can be convenient to calculate the essential matrix.
\begin{equation}
E=weg(pro,fpp),
\end{equation}
where weg($\cdot,\cdot$) is the process of using the weighted eight-point algorithm to calculate the essential matrix, E represents the computed essential matrix.

After the essential matrix is obtained, the label of the input point pair can be determined according to the symmetric epipolar distance~\cite{1} within a certain threshold. In this way, it is more accurate to estimate the camera pose than the one which directly divides the inlier and outlier labels.

\subsection{Correspondences Noise}

The process of calculating the descriptors and the coordinates of the keypoints may introduce errors. Generally, the descriptor only collects the local features around the keypoint, which only represents the parts information around the keypoints. Relying on partial information to match is bound to get inaccurate results. In many images, there are many areas with similar textures or similar feature descriptors, which may cause errors when matching keypoints in these areas. Due to the problem of calculation precision, the coordinates of keypoints will lead to the deviation of actual corresponding points. Both of these results introduce noise into subsequent processing, which is limited by existing processing methods.

The distance measure between the descriptors of keypoints is used to match the keypoints that are close to each other. Moreover, the distance measurement in feature matching only considers the similarity degree of descriptors on some attributes, such as Euclidean distance, Hamming distance, etc., and can not comprehensively measure the similarity between the descriptors. Therefore, noise is also introduced into the correspondences.


These mismatched correspondences can be regarded as matched point pairs that drift in the final result due to noise, as shown in Fig. 2. Or to say, the mismatched correspondences are the results of the superposition of the matched portion and the noise:
\begin{equation}
ol=ic\textcircled{+}nc,
\end{equation}
where \textcircled{+} represents the superposition of two components, ol represents an outlier, ic denotes the inlier component, and nc denotes the noise component.

In general, the noise component is a relatively small value compared with the inlier component, so we can filter the noise component by setting a threshold. For the features of different channels, the amplitude of the noise features to be filtered is different, so it is difficult to set the threshold artificially. We use the soft threshold module to make the network adaptively learn the threshold of filtering noise features of each channel.

\begin{figure}
\begin{center}
\centerline{\includegraphics[width=9cm]{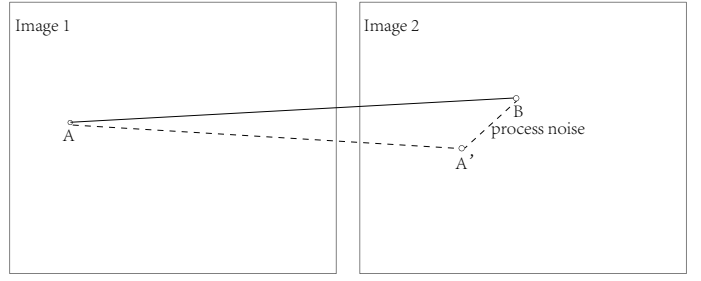}}
\end{center}
\caption{As can be seen from the figure, point A in image 1 actually corresponds to point A' in image 2. However, due to the noise introduced in the previous processing, the actual matching result of point A is changed, and it is matched to point B in image 2.}
\label{fig:short}
\end{figure}

\subsection{Soft Threshold Function}
Different soft threshold functions have different effects on the expression of the inlier components. Except for the linear soft threshold functions, we also use the quadratic function for the soft threshold function. Not only is the noise removed, but it also enhances the expression of inlier component features. So we get better results with a quadratic function. 

\subsection{Filtering Noise Block}
We design a filtering noise block for outlier rejection with soft threshold operation, as shown in Fig. 3. The specific processing flow of the filtering noise module is as follows.

\begin{figure}[ht]
\begin{center}
 \centerline{\includegraphics[width=9cm]{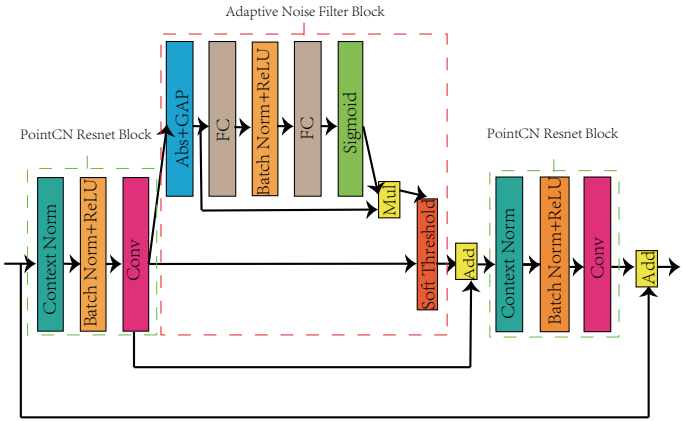}}
\end{center}
 \caption{Soft threshold filter noise module. We incorporated noise filtering into the PointCN Resnet Blocks. Conv represents convolutional layer, Abs denotes taking absolute value of features, Gap represents the global pooling of each channel, FC indicates full connection layer, Mul represents multiplication of corresponding positions, and Add denotes the addition of corresponding position of features.}
\label{fig:short}
\end{figure}

After the differentiable pooling, the ordered features first extracted the context information through the PointCN Resnet Block, then obtains the one-dimensional vector of channel width through taking the absolute value and global average pooling, 
\begin{equation}
f_c=Gap(Abs(f_i)),f_c\in R^{C\times 1\times 1},
\end{equation}
and then enters the two full connection layers a scale factor is obtained,
\begin{equation}
\lambda=FC(ReLU(BN(FC(f_c)))),\lambda \in R^{C\times 1\times 1},
\end{equation}
and the scale factor is normalized by a simoid function. The threshold value of filtered the noise features is obtained by multiplying the normalized result by the $f_i$.
\begin{equation}
t_s=Simoid(\lambda) \times f_i,t_s\in R^{C\times 1\times 1},
\end{equation} 
After the soft threshold processing, the noise-related features in the outliers are removed. 
\begin{equation}
f_o=stf(f_i,t_s),
\end{equation} 
where stf($\cdot$) represents the soft threshold function for filtering features generated by the noises.

Finally, the context information is further extracted after a PointCN ResNet Block, so that the influence of noise on the final result can be reduced while the context information is retained.

We add the soft threshold noise filter module to the PointCN Resnet Block and remove the noise extracted from the feature after obtaining the ordered cluster. It is a permutation invariant operation, that is, no matter the order of correspondences changes, the output corresponds to the input. This structure can self-adaptively determine the threshold of filtering features and filter each layer of channels with different thresholds, which can effectively remove the noise components in the outliers.

\section{Experiments}
\label{sec:typestyle}

\subsection{Experiments Setting}

\textbf {DataSets.} We evaluate our proposed method on the YFCC100M dataset~\cite{27}. The YFCC100M is an outdoor dataset, which contains 100 million images collected from the Internet. We use the results processed by~\cite{22}, 68 scenes are used for training and 4 scenes are used to verify the effect of our proposed method. 

\noindent \textbf {Evalution Metrics.} We use the angle differences between ground truth and the predicted results for rotation and translation as our evaluation criteria. Then the calculated mAP5(\%) is used as the main evaluation indicator.


\subsection{Comparisons with Other Baselines}
In this section, our proposed method is compared with other outlier rejection methods, RANSAC, PointCN, PointNet++, $N^3$Net, OANet. We report the results with RANSAC post-processing of all the methods in Table 1. Except for our method, all the other methods use the data of the paper~\cite{22}.

Our method not only preserves the context information between the input point pairs, but also filters out the noise features of the outliers, which reduce the output weights of the outliers, and relatively increases the output weights of the inliers, so our method achieves the best performance among these methods.

For a more comprehensive comparison, the precision, recall, and F-score of some methods in the dataset YFCC100M are listed in Table 2. On these indicators, our proposed method still achieves the best performance.



\begin{table}
\begin{center}
\begin{tabular}{cc}
\hline
Method & mAP5(\%)\\
\hline
RANSAC & 9.08\\
PointCN & 47.98\\
PointNet++ & 46.23\\
N$^{\rm 3}$Net&49.13\\
OANet&52.18\\
Ours &{\bf 52.63}\\
\hline
\end{tabular}
\end{center}
\caption{Comparison with the baselines on YFCC100M. mAP(\%) on are reported. All the Methods use SIFT to extract keypoints.}
\end{table}

\begin{table}
\begin{center}
\begin{tabular}{cccc}
\hline
Method & precision(\%)&recall(\%)&F-score \\
\hline
RANSAC & 41.83&57.08&48.28 \\
PointCN & 51.18&84.81&63.84\\
OANet&54.55&86.67&66.96\\
Ours &{\bf 55.24}&{\bf 87.18}&{\bf 67.6}\\
\hline
\end{tabular}
\end{center}
\caption{Result comparison of different methods. Symmetric epipolar distance is less than $10^{-4}$ is considered as an inlier.}
\end{table}





\section{Conclusion}
\label{sec:majhead}

We use the convolutional neural network based on the soft threshold function to filter the noise component in the outliers to obtain the relative pose between cameras. The outliers noise filtering network can reduce the noise component introduced in the previous processing. After filtering out the noise of the outliers, the weight of the outliers is closer to zero, and the outliers have less influence on the estimation of the camera pose so that the estimation of the camera pose can be more accurate. The experiments show that our proposed method achieves more performance than other methods.



\vfill\pagebreak

\bibliographystyle{IEEEbib}
\bibliography{strings,refs}

\end{document}